# Using Large Language Models for Automated Grading of Student Writing about Science

Chris Impey[1], Matthew Wenger[1], Nikhil Garuda[1], Shahriar Golchin[2], and Sarah Stamer[1]

(Corresponding Author: Chris Impey, cimpey@as.arizona.edu, ORCID 0000-0003-3515-167X)

## Abstract

Assessing writing in large classes for formal or informal learners presents a significant challenge. Consequently, most large classes, particularly in science, rely on objective assessment tools such as multiple-choice quizzes, which have a single correct answer. The rapid development of AI has introduced the possibility of using large language models (LLMs) to evaluate student writing. An experiment was conducted using GPT-4 to determine if machine learning methods based on LLMs can match or exceed the reliability of instructor grading in evaluating short writing assignments on topics in astronomy. The audience consisted of adult learners in three massive open online courses (MOOCs) offered through Coursera. One course was on astronomy, the second was on astrobiology, and the third was on the history and philosophy of astronomy. The results should also be applicable to non-science majors in university settings, where the content and modes of evaluation are similar. The data comprised answers from 120 students to 12 questions across the three courses. GPT-4 was provided with total grades, model answers, and rubrics from an instructor for all three courses. In addition to evaluating how reliably the LLM reproduced instructor grades, the LLM was also tasked with generating its own rubrics. Overall, the LLM was more reliable than peer grading, both in aggregate and by individual student, and approximately matched instructor grades for all three online courses. The implication is that LLMs may soon be used for automated, reliable, and scalable grading of student science writing.

## Acknowledgements

We acknowledge fruitful conversations with Sanlyn Buxner about the statistical analysis and the educational implications of this research, and with Alexander Danehy on the computational basis for the analysis and on the strengths and weaknesses of the current generation of large language models. This work was supported in part by a grant from the National Science Foundation, award DUE-2020784.

## Author Contributions

All authors contributed to the design and conception of the study. Evaluation material for the MOOCs was created by Matthew Wenger, who also acted as the instructor grader for this project. Implementation of the LLMs for grading student writing was done by Nikhil Garuda and Shahriar Golchin. The statistical analysis was done by Nikhil Garuda. The gathering of peer grading information was carried out by Nikhil Garuda and Sarah Stamer. The first draft of the manuscript was written entirely by Chris Impey. All authors commented on successive versions of the manuscript and all authors read and approved the final manuscript.

---

[1] Department of Astronomy, University of Arizona, Tucson, AZ 85721, United States
[2] Department of Computer Science, University of Arizona, Tucson, AZ 85721, United States

## Introduction

Artificial intelligence (AI) has had a profound effect on diverse fields (Ryan, 2023; Shamshiri et al., 2024; Thiranavukasaru et al., 2023). AI has also impacted education at every level (Zhang & Aslan, 2021). College leaders see both promise and peril in this disruptive technology. In one survey of college leaders' opinions about Generative AI, 78% agreed that the tools offer an opportunity to improve how colleges educate, operate, and conduct research, but 57% also thought that the same tools pose a threat to how colleges educate, operate, and conduct research (Anft, 2023). Recently, large language models (LLMs) and tools such as ChatGPT (Ouyang et al., 2022) have shown great potential to help students learn but have also led to concerns about plagiarism and a degradation in the ability of students to write and synthesize information (Grassini, 2023). A complex typology of AI's capabilities affects every aspect of education, from tutoring and assessment to the way institutions admit students and identify those who are at risk (Holmes & Tuomi, 2022). The literature discussing LLMs in the classroom has mostly focused on instructors and students using them to generate educational content (Kasneci et al., 2023). In 2023, Khan Academy and OpenAI announced a partnership using GPT-4 as a learning assistant tool to facilitate student learning (OpenAI, 2023; Khan, 2023). Students will be able to ask questions about content as they would an instructor. More sophisticated AI assistants for students (and instructors) are anticipated in the future.

We embarked on a project to see if LLMs could be useful in massive open online courses, or MOOCs. MOOCs are typically free and are open to anyone in the world who has access to a computer and the Internet. In their first ten years, MOOCs have grown to nearly 20,000 courses, offered by 950 universities, and serving 220 million students worldwide (Shah, 2021). MOOCs are of interest to researchers because they are an informal learning environment where people can learn about science without enrolling in a university class, particularly adult learners (Falk and Needham, 2013). Although MOOCs resemble formal classes, with video lectures, quizzes, and activities, the learning is in a self-directed environment guided by individual needs and interests (Oakley & Sejnowski, 2019). Unlike the college setting, learners do not get grades or transferable credit and classes do not typically contribute to a degree program. MOOCs have an international audience that encompasses many developing countries, so they can play an important role in the democratization of education (Impey, 2020). The current study builds on prior work examining peer assessment in MOOCs. Although Formanek et al. (2017) showed that participation in peer grading is correlated with student engagement and course completion, it has also been found that peer grading can be inconsistent and there are problems with reliability and validity (Formanek et al., 2019; Gamage, Staubitz, & Whiting, 2021; Usher & Barak, 2018; Yousef, & Sumner, 2021).

This study investigates whether an existing top performing LLM can equal or surpass peer grading in reliability relative to an instructor (Bojic, Kovacevic, & Caparkapa, 2023). Instructor model answers are one input, but another is an instructor-generated rubric, since predicting rubric scores has been found to be essential to automated essay grading (Kumar & Boulanger, 2020). We also investigate whether LLMs can be effective at generating grading rubrics. This can be useful for pre-existing writing assignments where no rubric is provided. If LLMs can approach instructor reliability, they can potentially be used in low-stakes MOOCs with tens of thousands of learners, where grading by peers is a burden and grading by a human instructor is essentially impossible.

Among the three courses included in this study, writing assignments for introductory astronomy and astrobiology are the easiest for a human or a machine to evaluate because the assignments are content based. The course on the history and philosophy of astronomy is more challenging because answers to some of the questions depend on speculation or hypothetical situations, where the judgement can be subjective (Golchin et al. 2024, Impey, 2023).

We address the following research questions: can the LLMs (1) generate a grade comparable to that of an instructor, (2) match or exceed the reliability of peer grading, and (3) create a grading rubric that will produce LLM grades comparable to that of an instructor? This work is intended to apply to writing assignments on a variety of topics and act as a proof-of-concept to explore whether LLMs can reliably evaluate science writing and whether automated grading can be scaled to many thousands of students in online classes.

## Previous Work

Keeping online learners motivated and involved is far more difficult than in a face-to-face class (Martin & Borup, 2022). MOOCs present a particular challenge since there is little opportunity for direct, real-time interaction with other students or the instructor. Prior research shows that MOOC completion rates are persistently low (Jung & Lee 2018; Onah, Sinclair, & Boyatt, 2014; Reich & Ruipérez-Valiente, 2019; Wang & Baker, 2015). Course design and pedagogy can play a role in improving these results. Active learning through lab activities and writing assignments has been shown to increase MOOC engagement and completion rates (Formanek et al., 2019; Shah et al., 2022). In addition to being connected with course engagement and completion, writing assignments promote learning. Suitably designed writing assignments can enhance science literacy (Hand et al., 1999) and can be an effective way to improve student reasoning skills and critical thinking (Graham, Kiuhara, & MacKay, 2020; Libarkin & Ording, 2012; Quitadamo & Kurtz, 2007). Unfortunately, assessing, grading, and providing feedback on writing assignments in large classes is notoriously difficult. At the scale of MOOCs, it is impractical for instructors to grade thousands of writing assignments.

The current solution to the problem of scale is peer-grading. In the Coursera system, every writing assignment is graded by multiple other randomly selected students in the same course, using an instructor-provided rubric. Students are allowed to grade peer assignments only after they have submitted their own, and final grades are determined by taking the median of the peer grades. Although peer-grading does have some positive effects, such as improving learner engagement and motivation, as a grading methodology it often has limited reliability and validity (Formanek et al., 2017; Formanek et al., 2019; Gamage et al., 2021; Usher & Barak, 2018; Yousef, & Sumner, 2021). A previous study analyzed peer grading for the astronomy course used here (Formanek et al., 2017). In that study, using data from 2015, peer review for 300 assignments out of 4 points had a mean score of 3.39, slightly lower than the instructor mean score, with a standard deviation of 0.78. In terms of reliability, unsurprisingly, instructor grades were the most reliable, followed by trained undergraduate graders, followed by peer graders. The correlation between the median instructor grades and the median peer grades was moderate ($r = 0.49$).

Automatic grading of writing assignments using machine learning is another possible solution to the problem of grading at scale. There is evidence that MOOCs using automated grading systems have higher completion rates (Kruchinin, 2019). AI techniques are beginning to be used in MOOCs since they can readily be scaled for many thousands of learners. The use of machine learning to grade or assist with the grading of short answers has been attempted before using a variety of methods (Borad & Netak, 2021; Leacock & Chodorow, 2003; Weegar & Idestam-Almquist, 2024). A literature review in 2020 found twenty papers using AI for the assessment of students (Sanchez-Prieto et al., 2020). Four analyzed student behaviors, six investigated student feelings or sentiments, and ten assessed student achievement through AI-based methods. Among the ten, the focus was grading multiple-choice tests, lab exercises, concept maps, and short-answer questions. None investigated longer writing assignments, as we do in this work. Recently, GPT-3.5 (Ye et al., 2023) has been shown to have an accuracy of 65-95% for grading multiple-choice tests across ten different science topics (Alseddiqi et al., 2023). This variability in performance can be attributed to potential data contamination issues in LLMs or their inherent probabilistic behavior during text generation (Golchin & Surdeanu, 2023a, b). One study has used LLMs to validate peer-assigned essay scores in a Coursera MOOC (Morris et al., 2023). In a recent study, the BERT language model was effective in evaluating writing based on grammar, semantics, coherence, and prompt relevance (Vanga et al., 2023). Another study using the RoBERTa language model shows that this language model could outperform human raters (Beseiso et al., 2021).

## Methods

### *Research Context*

This research uses data gathered from three MOOCs. Our education research group has been offering MOOCs through Coursera since 2013 (Impey, Wenger, & Austin, 2015). The classes utilized in this study are on the topics of astronomy (Impey et al., 2016), astrobiology (Impey, Wenger, & Riabokin, 2023), and the history and philosophy of astronomy (Impey, 2023). Respectively, they are titled "Astronomy: Exploring Time and Space," "Astrobiology: Exploring Other Worlds," and "Knowing the Universe: History and Philosophy of Astronomy." Together, they have enrolled nearly 225,000 learners in 190 countries. All three courses share a similar format: weekly modules that contain both video lecture content and supplemental documents including lecture slides. The assessments include automatically graded multiple-choice quizzes using the quiz system provided by Coursera as well as peer-graded writing assignments. These classes are available for free and are aimed at learners who have no background in science. They are designed for a high-school or introductory college-level audience.

The first course, "Astronomy: Exploring Time and Space," was launched in 2013. It is an introductory-level survey of astronomy that covers topics including history, the night sky, telescopes, light, the Solar System, stars, galaxies, and life in the universe. It is the longest course with 11 modules that contain 18 hours of video lectures and lecture slides, as well as optional supplementary materials such as articles about relevant topics and podcast episodes. Assessments include quizzes, five peer-graded writing assignments at the end of modules 2, 4, 7, 8, and 10, and a project at the end of week 9. There are currently 194,032 learners enrolled, and 9,655 have completed the course.

The second course, "Astrobiology: Exploring Other Worlds," was launched in 2019 and is an introductory-level course that covers topics including planet formation, exoplanet detection, habitability, life, biology, and the search for extraterrestrial intelligence. It has six modules that contain 10 hours of video lectures and lecture slides. Assessments include short quizzes after each lecture video, three peer-graded writing assignments at the end of modules 2, 3, and 5, and a final written and peer-graded project at the end of week 6. There are currently 26,337 learners enrolled, and 1,302 have completed the course.

The third course, "Knowing the Universe: History and Philosophy of Astronomy," was launched in 2022. It is an introductory-level course that covers topics including prehistory, the advent of science, world cultures and the development of observational astronomy, the development of modern science, modern physics and relativity, cosmology, and life beyond Earth. It has eight modules that contain 18 hours of video lectures and lecture slides. Assessments include quizzes and seven short peer-graded writing assignments at the end of modules 2, 3, 4, 5, 6, 7, and 8. Each writing assignment has three question options from which students must select one. There are currently 3,916 learners enrolled, and 181 have completed the course.

The writing assignments in each class were designed independently and emphasized slightly different criteria. The questions for Introductory Astronomy and Astrobiology focused on applying and analyzing course materials and responding with accurate scientific answers. The questions for the History and Philosophy course are more open-ended and allow students more flexibility in how to answer. Writing assignments for all three courses asked students to organize their thoughts, state their ideas clearly, and to use supporting evidence. Students are provided with instructions that include a writing prompt and the criteria that will be used to grade the assignments. These criteria were then used in the grading rubrics for scoring peer assignments. Rubrics were designed with non-expert peer-graders in mind and included several features which have been found to improve the consistency of scoring (Jonsson & Svingby, 2007). The rubrics for these courses are generally analytical and the Introductory Astronomy and Astrobiology questions are topic specific while those for the History and Philosophy class are more general to account for a variety of topics and responses. For more subjective criteria (organization, clarity) the scale was kept small in an attempt to produce more consistent scores. Often these criteria are scored as present or not, with little requirement for graders to judge the level of each criterion. To the extent possible, a detailed breakdown for how points should be allocated was provided in the rubric. The text for all writing assignment questions and grading rubrics used in this research study are provided in Appendix A. All three courses use the same built-in Coursera peer-grading system.

This research was overseen by the Institutional Review Board at our university. Students in these courses completed an optional survey at the beginning of each class that includes questions about demographics as well as their consent to participate, or not, in our research. Data was gathered only from students who both completed the survey and agreed to participate in our research study. All answers were de-identified prior to instructor and LLM grading. While attempts were made to reduce bias in the sampling process, we acknowledge that these results are likely not to be representative of the global population. They are, however, representative of Coursera learners. Most learners on Coursera are between the ages of 24 and 44 and a majority of students in our classes are male. The largest difference in gender was in the History and Philosophy course (34% female vs. 65% male) and the smallest difference was in the Astrobiology course (45% female, 54% male). Most of the students in our courses live in either the United States or India. Since all data are de-identified, sample selection was minimally influenced by these characteristics.

### Research Data

In order to answer the research questions in this study, we needed a sample of student writing assignments and three sets of grades for those assignments: 1) Instructor grades, 2) Peer Grades, and 3) GPT-4 Grades. The first step was to select the assignments. To have a sample large enough to conduct a statistical analysis, yet possible for a single instructor to grade in a reasonable timeframe and with available LLM resources, we aimed for a sample size of around 100 assignments. Although analyzing grades for a single question might increase statistical power, it does not show the generalizability of using an LLM across different types of courses and questions. It also opens up the possibility of cherry-picking an assignment that would produce an unrepresentative result, either because of the structure of the question, or the construction of the grading rubric. By selecting responses to multiple different questions from different courses, these risks are reduced, the data selection process is more fair, the data is more representative of the variety of questions and rubrics used across these courses, and the generalizability of the result is improved. With this in mind, we gathered a subset of answers from 120 learners to each of 12 questions across the three courses, with 10 answers for each assignment. Answers were chosen from five questions in the Introductory Astronomy class, three questions from the Astrobiology class, and four questions from the History and Philosophy of Astronomy class. The assignments in the Introductory Astronomy class were worth 9 points each, except for the first assignment which was worth 6 points. Astrobiology class assignments were worth 10 points each, and the assignments in the History and Philosophy of Astronomy class were worth 4 points each. For the introductory astronomy and astrobiology courses, the students were asked to write a response of 250 to 750 words, and for the history and philosophy class, the instructions asked for 250 to 300 words. These word limits were not strictly enforced, so there is some variation.

The de-identified assignments were purposefully sampled (Ramos Rojas et al., 2017) to have a spread of peer grades that represent a full range of possible scores. Answers were chosen blindly without examining the content. This selection method was chosen because a majority of students in each class received full marks on the assignments and initial attempts to select assignments randomly resulted in a highly skewed dataset with mostly high scores, which would have provided no dynamic range. Another data selection challenge arose from the fact that the history and philosophy course was launched most recently and has a lower enrollment. As a result, fewer assignments were available. Because each writing assignment had three question options for students to choose from, the assignments were first examined to determine the topic covered and which questions had the most responses. The questions with the most responses were selected for analysis. As with the other two courses, a sample of assignments was selected to represent as much of the range of scores as possible, given the limited available answers.

To answer the first research question, one of the instructors graded the student assignments using the rubrics he had created (Stevens and Levi, 2012; Pisano et al., 2021). These are the same rubrics that are available to students and peer-graders in each course. The grading was made blind by having one of the authors provide writing assignments to the instructor without grade information. The instructor then graded the assignments with part scores according to the rubric.

A full set of the questions and grading rubrics used in this study have been provided in Appendix A. These same rubrics were provided to the LLMs for this experiment. Example or model answers (also referred to as instructor-provided answers) were written by the instructor to represent an ideal version of an acceptable correct answer for each assignment. Model answers are available upon request by contacting the authors.

Grades from GPT-4 were generated by providing information to each model using the template prompts shown in Figures 4, 5, and 6 in Appendix C. The three conditions used in the prompts included: Prompt 1: instructor-provided answer only; Prompt 2: instructor-provided answer plus instructor rubric; and Prompt 3: LLM (AI) generated rubric plus instructor answers. Example rubrics generated by GPT-4 can be found in Appendix B.

For the second research question, peer grades were downloaded from Coursera. Learners in these courses are required to grade assignments of their peers to complete the course. They are supposed to review three or four assignments, or their final grade for the assignment is reduced by 20%. The final writing assignment grade is the median of the scores received from the peer graders.

To answer research question 3, LLM rubrics were generated by providing information about the category of the course, the audience, and the total grades and breakdown along with the instructor model answer. Example GPT-4 generated rubrics are provided in Appendix B. These LLM rubrics were then provided to GPT-4 in a prompt to grade student answers. Example prompts are provided in Appendix C.

## Results

In this study, we addressed the following research questions: can the LLMs (1) generate a grade comparable to that of an instructor, (2) match or exceed the reliability of peer grading, and (3) create a grading rubric that will produce LLM grades comparable to that of an instructor?

We normalized the scores by converting them to percentages and conducted a paired-test analysis of the data to determine whether we could identify differences between the instructor grades, peer grades, and GPT-4 grades. Due to the skewed distributions of the scores we obtained (shown below in Figure 1), we wanted to check the assumptions of normality and homogeneity of variances.

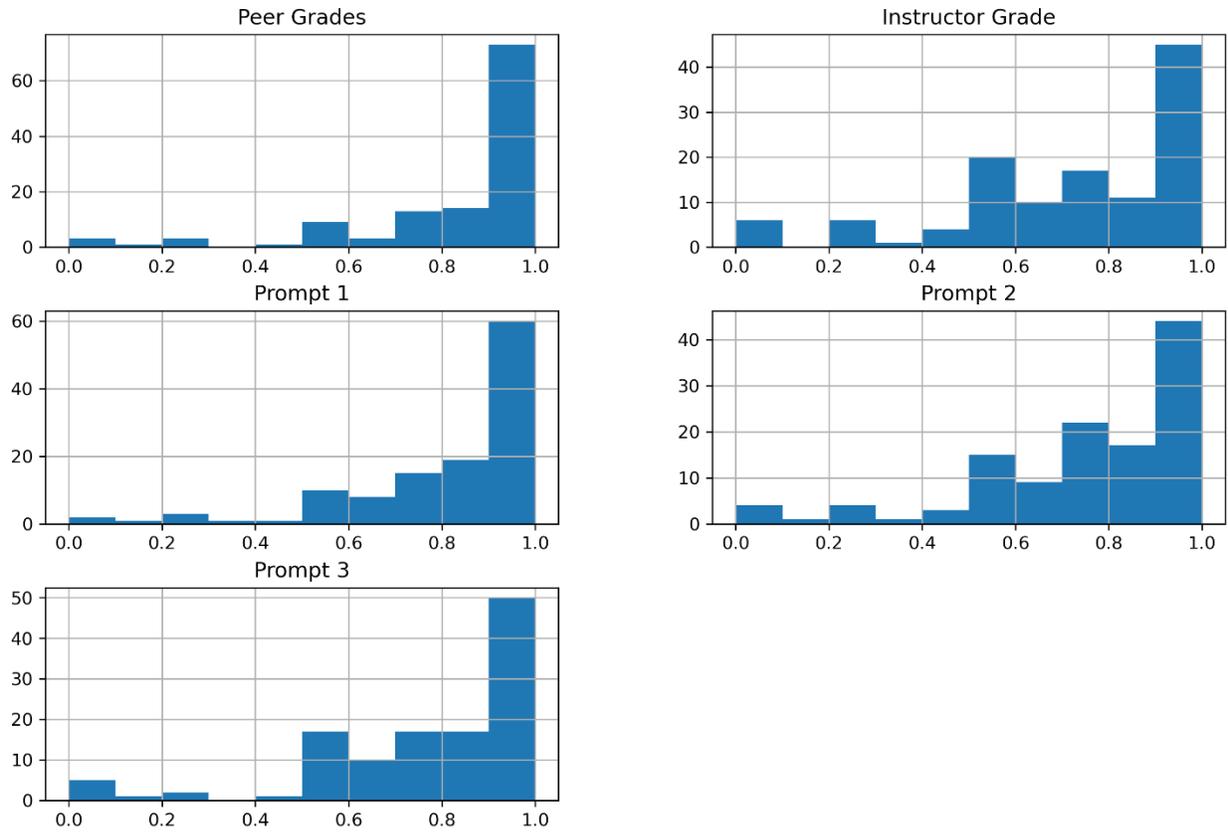

**Fig. 1** Histograms of normalized grades from instructor, peers, and LLM prompts for all responses showing skewed distribution of scores.

Due to violations of normality as indicated by Shapiro-Wilk (p=0.003) (Shapiro & Wilk, 1965; Razali & Wah, 2011) and homogeneity of variances indicated by Levene's tests (p=0.021) (Levene, 1960; Brown & Forsythe, 1974), non-parametric statistical methods were employed. Specifically, the Friedman Test (Friedman, 1937; Friedman, 1940) was used to assess differences among graders, and Conover's post-hoc tests (Conover, & Iman 1979; Conover, 1999) were conducted for pairwise comparisons. A Friedman Test was conducted to compare the grades among the Instructor, Peer, and LLM graders. The results from a post-hoc pairwise comparison using Conover's test with Bonferroni correction (Pereira, Afonso, & Medeiros, 2015), to reduce false positives, are presented below in Table 1. For the LLM, Table 1 includes three lines. The first line is the average score determined by the LLM when provided with the instructor's model answer (Prompt 1). The second line is the average score from the LLM when provided with the instructor's model answer and the rubric written by the instructor (Prompt 2). The third line is the average score using a rubric created by the LLM based on the instructor's model answer (Prompt 3).

Table 1. Post-Hoc Analysis of Friedman Test

| Model | | | Instructor Grade | Peer Grades | Prompt 1: Instructor Answer | Prompt 2: Instructor Answer + Rubric | Prompt 3: AI Rubric + Instructor Answer |
|---|---|---|---|---|---|---|---|
| | | Prompt | | | | GPT-4 | |
| | | Instructor Grade | 1.00 | ~0.00* | ~0.00* | 1.00 | 1.00 |
| | | Peer Grades | ~0.00* | 1.00 | 1.00 | ~0.00* | ~0.00* |
| GPT-4 | | Prompt 1: Instructor Answer | ~0.00* | 1.00 | 1.00 | ~0.00* | 0.001* |
| | | Prompt 2: Instructor Answer + Rubric | 1.00 | ~0.00* | ~0.00* | 1.00 | 1.00 |
| | | Prompt 3: AI Rubric + Instructor Answer | 1.00 | ~0.00* | 0.001* | 1.00 | 1.00 |

**Table 1**  Results of a post-hoc analysis for the Friedman chi-square analysis using Conover's test with Bonferroni correction. An asterisk * indicates results that are significant at $p < 0.05$.

The results indicate significant differences between the instructor's grades and both peer grades ($p < 0.001$) and the LLM when GPT-4 is only provided with an example answer ($p < 0.001$). No significant differences were found between the instructor's grades and LLM grades when the GPT-4 is prompted with both an example answer and a rubric, whether it is instructor provided, or LLM generated. ($p = 1.000$). LLM grades with only an example answer are not statistically different than peer grades.

Because the types of questions and rubrics are very different between courses, we wanted to look more closely to see if there were differences in the way that GPT-4 graded the questions. We ran a similar paired-test analysis for each class, but the number of assignments was too small and none of the differences between graders was significant.

Instead, we used descriptive statistics to examine the results for individual questions. The results of this analysis are summarized in Table 2, which shows the average scores for each question in each course. One-sigma standard deviations for each average were determined from bootstrap resampling of the scores (Efron, 1979). To mimic the situation of a MOOC setting, 10,000 bootstrap samples were taken in each case. The p-values for these results are presented below in Table 3. All are above 0.05, implying that there is no statistically significant difference in average grades.

Table 2.  Instructor grades, Peer Grades, and LLM Grades

| Model | Prompt | Courses | | | | | | | | | | | |
|---|---|---|---|---|---|---|---|---|---|---|---|---|---|
| | | Astronomy | | | | | Astrobiology | | | History and Philosophy | | | |
| | | Q1 | Q2 | Q3 | Q4 | Q5 | Q1 | Q2 | Q3 | Q1 | Q2 | Q3 | Q4 |
| | Instructor Grades | 3.90 ± 0.54 | 8.2 ± 0.37 | 7.51 ± 0.92 | 7.41 ± 0.86 | 5.51 ± 0.94 | 6.8 ± 1.09 | 6.7 ± 0.83 | 7.89 ± 0.85 | 3.50 ± 0.21 | 2.39 ± 0.29 | 2.70 ± 0.20 | 2.40 ± 0.15 |
| | Peer Grades | 5.15 ± 0.27 | 7.55 ± 0.75 | 7.41 ± 0.82 | 7.45 ± 0.46 | 7.40 ± 0.83 | 7.50 ± 0.79 | 7.46 ± 1.02 | 9.04 ± 0.52 | 3.60 ± 0.21 | 3.69 ± 0.20 | 3.40 ± 0.21 | 3.80 ± 0.19 |
| GPT-4 | Prompt 1: Instructor Provided Answer | 4.75 ± 0.41 | 8.65 ± 0.20 | 7.61 ± 0.87 | 7.51 ± 0.77 | 6.21 ± 0.90 | 7.50 ± 0.70 | 7.91 ± 0.68 | 8.10 ± 0.33 | 3.50 ± 0.21 | 3.25 ± 0.23 | 3.65 ± 0.14 | 3.2 ± 0.24 |
| | Prompt 2: Instructor Provided Answer + Rubric | 4.40 ± 0.41 | 8.30 ± 0.28 | 7.31 ± 0.86 | 6.91 ± 0.90 | 5.91 ± 1.06 | 7.11 ± 0.95 | 7.41 ± 0.83 | 7.50 ± 0.41 | 3.20 ± 0.31 | 3.10 ± 0.17 | 3.20 ± 0.19 | 2.70 ± 0.28 |
| | Prompt 3: AI Rubric + Instructor Answers | 4.40 ± 0.48 | 8.50 ± 0.27 | 7.61 ± 0.87 | 7.06 ± 0.89 | 6.36 ± 1.10 | 7.11 ± 0.85 | 7.11 ± 0.98 | 7.50 ± 0.29 | 3.20 ± 0.22 | 2.95 ± 0.23 | 3.20 ± 0.18 | 2.95 ± 0.23 |

**Table 2**  Average grades from 10 students' writing assignments for each question in each course. Results from GPT-4 are displayed for each prompt. One-sigma standard deviations (shown below the averages) for each average were determined from bootstrap resampling of the scores.

Table 3. p-values for results in Table 1

| Model | Prompt | Courses | | | | | | | | | | | |
|---|---|---|---|---|---|---|---|---|---|---|---|---|---|
| | | Astronomy | | | | | Astrobiology | | | History and Philosophy | | | |
| | | Q1 | Q2 | Q3 | Q4 | Q5 | Q1 | Q2 | Q3 | Q1 | Q2 | Q3 | Q4 |
| | Peer Grades | 0.48 | 0.52 | 0.95 | 0.98 | 0.50 | 0.66 | 0.65 | 0.50 | 0.83 | 0.50 | 0.58 | 0.61 |
| GPT-4 | Prompt 1: Instructor Provided Answer | 0.49 | 0.49 | 0.97 | 0.96 | 0.66 | 0.65 | 0.53 | 0.87 | 1.00 | 0.52 | 0.55 | 0.51 |
| | Prompt 2: Instructor Provided Answer + Rubric | 0.52 | 0.78 | 0.89 | 0.74 | 0.79 | 0.87 | 0.64 | 0.71 | 0.61 | 0.52 | 0.55 | 0.55 |
| | Prompt 3: AI Rubric + Instructor Answers | 0.55 | 0.60 | 0.95 | 0.79 | 0.64 | 0.86 | 0.80 | 0.72 | 0.57 | 0.51 | 0.55 | 0.51 |

**Table 3**  This table shows the p-values for the differences in the average scores of 10 students' writing assignments for each question. In Table 2, there is no row for instructor grades since that acts as the baseline for the p-values, but we instead show the p-values obtained from the peer grades. These p-values were computed through bootstrap resampling (Efron, 1979) with 10,000 iterations. (Adapted from Golchin et al., 2024)

We also calculated root mean square (RMS) values to examine the agreement of instructor and LLM scores. The RMS results after averaging over all questions in each class are shown in Table 3.

Table 4. RMS Differences between Instructor and LLM Grades

| Model | Prompt | Astronomy | Astrobiology | History and Philosophy |
|-------|--------|-----------|--------------|------------------------|
| GPT-4 | Prompt 1: Instructor Answer | 0.54 | 0.82 | 0.76 |
| | Prompt 2: Instructor Answer + Rubric | 0.38 | 0.50 | 0.48 |
| | Prompt 3: AI Rubric + Instructor Answer | 0.49 | 0.37 | 0.49 |

**Table 4** This table shows the root mean square (RMS) differences between instructor grades and the grades assigned by GPT-4 across three courses: Astronomy, Astrobiology, and History and Philosophy of Astronomy. The results indicate that GPT-4 generally achieves close alignment with instructor grades, particularly when an instructor-provided rubric is used. The smallest RMS differences for GPT-4 occur when using the instructor answer plus rubric, suggesting that this combination provides the most accurate grading relative to the instructor.

### Large Language Model vs. Instructor

For research question 1: Can the LLM generate grades comparable to those of an instructor? The answer is yes, when given appropriate information, GPT-4 was able to produce grades that are not statistically different than the instructor. These results, however, depend on how the LLM prompt is constructed, and what information is provided. Based on the results in Table 1, the LLM produces grades that are significantly different from the instructor when only given an example answer. The best performance is observed for GPT-4 when the prompt includes both an example answer and a grading rubric, whether it is instructor-provided, or LLM generated, the LLM produced grades that are not significantly different from the Instructor's grades, suggesting that the LLM can effectively replicate instructor grading when it is provided with sufficient information.

This result seems reasonable since the 'Instructor Answer + Rubric' prompt aligns closely with what the instructor uses when grading. Incorporation of an AI-generated rubric has the potential to create more robust rubrics by leveraging correlations from instructor answers. It's important to note, however, that relying solely on correct answers alone may not be sufficient, and the use of a grading rubric is crucial for ensuring greater reliability in the grading process.

### Individual Courses

Although the paired-test results failed to detect significant differences between graders by class due to the small numbers and lack of statistical power, some variations in the descriptive statistics are notable. For the class "Astronomy: Exploring Time and Space," The LLM consistently graded higher than the instructor across all prompts, Notably, the LLM struggled to align with the instructor's grades in the first question due to its open-endedness. On the second question, there is improved agreement with the instructor, as both LLMs tend to grade higher. On questions three and four GPT-4 produced grades similar to the instructor. The fifth question has the lowest instructor scores, GPT-4 graded substantially higher than the instructor. It is noteworthy that the results remain consistent whether the instructor's answer is paired with an AI-generated rubric or with the instructor's rubric.

In "Astrobiology: Exploring Other Worlds," the questions allow students to give both open-ended answers, allowing for a thorough exploration of the topics, and well-structured answers, presenting information in a logical and organized fashion. On the first question, GPT-4 consistently scores higher than the instructor for all prompts. The best results were obtained when the LLM was provided with an instructor's model answer and rubric. On the second question, the pattern is similar, with GPT-4 being more lenient in its scoring. On the third question, GPT-4 had excellent agreement with the instructor.

The "Knowing the Universe: History and Philosophy of Astronomy" course presents the greatest challenge for either an instructor or for an LLM because the subject matter is very broad and involves conceptually challenging material. An answer might be based as much on plausible speculation as on a consensus among experts. Students are therefore challenged to use higher-order thinking skills rather than mere fact retrieval. Instructor's scores were lower as a percentage than for the other two courses. On the first question, the LLM scored lower than the instructor by a substantial amount, with the best agreement for a prompt with the AI-generated rubric. On the second question, the agreement is generally better. On the third question, the agreement is poor, with three of the prompts several standard deviations away from instructor scores. On the fourth question, the agreement is again poor, with the best result obtained using the instructor model answer along with an AI-generated rubric.

### *Reliability of Large Language Models and Peer Grading*

For research question 2: Can the LLM match or exceed the reliability of peer/instructor grading? The answer is yes, The LLM's reliability matches or exceeds that of peer grading. In line with our previous research (Formanek et al., 2017), this study confirmed that peer grades differed significantly from the instructor's grades, indicating variability in peer assessments. GPT-4 had a similar problem with reliability when provided with only an example answer. Table 1 shows that GPT-4 produced grades that differ significantly from the instructor when only prompted with an example answer, yet these results were not statistically different from peer graders. This difference disappears when GPT-4 was prompted with both a rubric and an example answer. This highlights the importance of prompt design in LLM grading performance and demonstrates that GPT-4 is capable of providing grades that are more reliable than peer grades and more closely aligned with instructor grades. The Intraclass Correlation Coefficient (ICC) was calculated to assess inter-rater reliability (Shrout & Fleiss, 1979). An ICC of 0.92 (95% CI: 0.89 - 0.94) was obtained, indicating excellent reliability among the graders. The high ICC value (0.92) demonstrates that the LLM's grading is consistent and reliable, matching or exceeding the reliability of peer grading.

These results are also visible in the mean scores for each question in Table 2. Peer graders generally give higher scores and are more lenient than the instructor in almost every case. This means that LLMs are as capable of accurately grading student writing assignments as the peer-graders currently used in these MOOCs. LLMs clearly have the potential to act as proxies for the instructor, avoiding some of the pitfalls and limitations of using novices (other students in the class) to grade student assignments

### *Language Model Rubrics*

Research question 3 was addressed by prompting GPT-4 to create a rubric based on the question text and the example answer provided by the instructor. These rubrics were evaluated by having the LLM assign grades using the AI-generated rubric in combination with the instructor-provided answer, and then comparing that with the results of the instructor rubric and instructor-provided answer. The differences between the resulting scores were not statistically significant, indicating that LLM-generated rubrics are of similar utility to the instructor-provided rubrics in terms of using them for this automated grading procedure. They also improve the results relative to providing the instructor answer alone, bringing grade agreement in line with using the instructor rubric.

### Comparisons for Individual Students

The comparisons just described are averages across all ten students sampled. However, a student cares about the accuracy of their own grade more than the class average. To address this, we made a direct comparison of instructor, GPT-4 (Prompt 2: Instructor Provided Answer + Rubric), and peer-graded scores on all twelve assignments for each of the ten students individually. Prompt 2 was chosen for the GPT-4 scores because it is the most direct comparison, as it uses the same information available to instructors and peers. Additionally, we investigated the dispersion among the peer grades, where there were four for each assignment in the introductory course, and three for each assignment in the astrobiology and history and philosophy courses. Figure 2 is a scatterplot of the difference in grades between the instructor and the median for the peer grades on the x-axis and the difference in grades between the instructor and GPT-4 on the y-axis. Histograms are also shown.

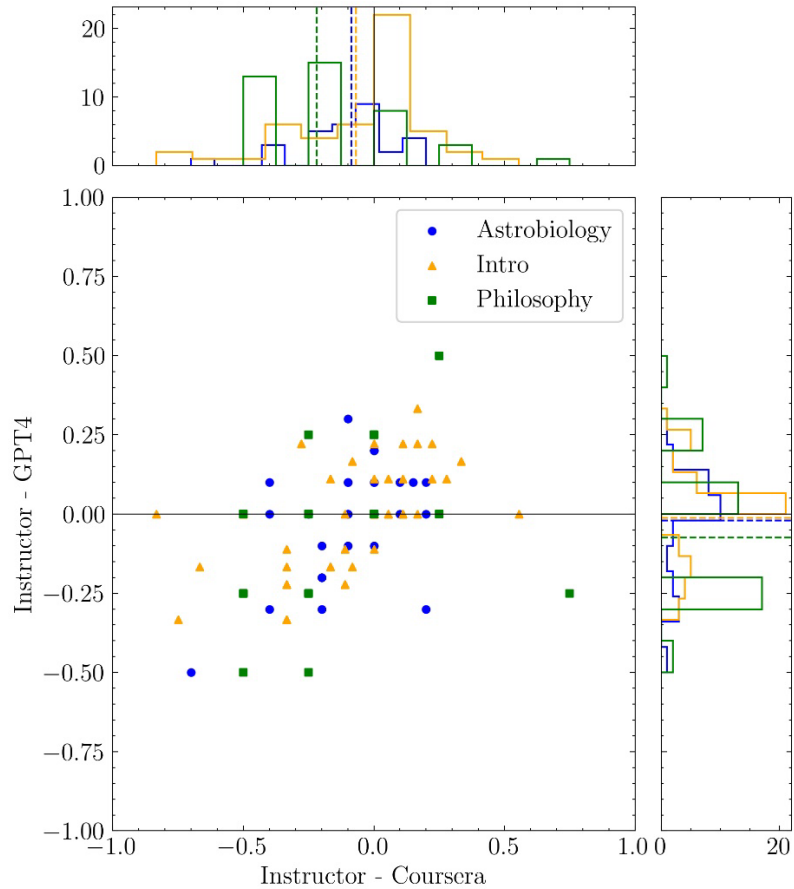

**Fig. 2** Scatter plot showing the difference between the grades by the instructor and the median of the peer scores received by Coursera for each student on the x-axis, and the difference between the grades by the instructor and the LLM Model (GPT-4 prompt 2) on the y-axis. Symbols for the three courses are yellow triangles (Introductory Astronomy), blue circles (Astrobiology), and green squares (History and Philosophy of Astronomy). The two histograms show the distribution of the data points for these differences along each axis. Dashed lines in the histograms indicate the means of the three classes for the two measures of grade difference.

For the introductory course, the mean instructor minus GPT-4 score is -0.01 ± 0.15, compared to the mean instructor minus peer grading score of -0.07 ± 0.26. The LLM grade is very close to the instructor grade, with a smaller dispersion than the peer grade. For the astrobiology course, the mean instructor minus GPT-4 score is -0.02 ± 0.17, compared to the mean instructor minus peer grading score of -0.09 ± 0.20. The LLM grade is very close to the instructor grade, with a slightly smaller dispersion than the peer grade. Lastly, for the history and philosophy course, the mean instructor minus GPT-4 score is -0.08 ± 0.22, compared to the mean instructor minus peer grading score of -0.22 ± 0.27. The LLM grade is much closer to the instructor grade and has a slightly smaller dispersion than the peer grade.

In one notable anomaly from the astrobiology course, both peer grades and GPT-4 gave substantially higher scores than the instructor. Upon closer examination, this answer was exceptionally long and contained rambling passages full of extraneous information. GPT-4 clearly struggled with this passage, and peer graders gave full credit even though the instructor recognized that the answer was largely incorrect.

Overall, LLM grades and peer grades are always more lenient than instructor grades, with a few examples in the astrobiology class where peer graders were considerably more lenient than either the instructor or GPT-4. Overall, GPT-4 performs better than peer grading.

To further understand why there was a higher discrepancy among peer grades and how that discrepancy compared to the instructor, we retrieved from Coursera the median peer grades for each student to see the dispersion of peer grades for each question. We calculated a representative score by taking the average of the 10 median student grades and converting them into percentages. Then, we calculated the mean absolute deviation for each of the peer grades and used the average of those deviations for the error bars on each score. For the instructor scores, we used the average of the instructor-given scores for the 10 students, converted into percentages, and the average of the mean absolute deviation for each score as the error bars. The results are shown below in Figure 3.

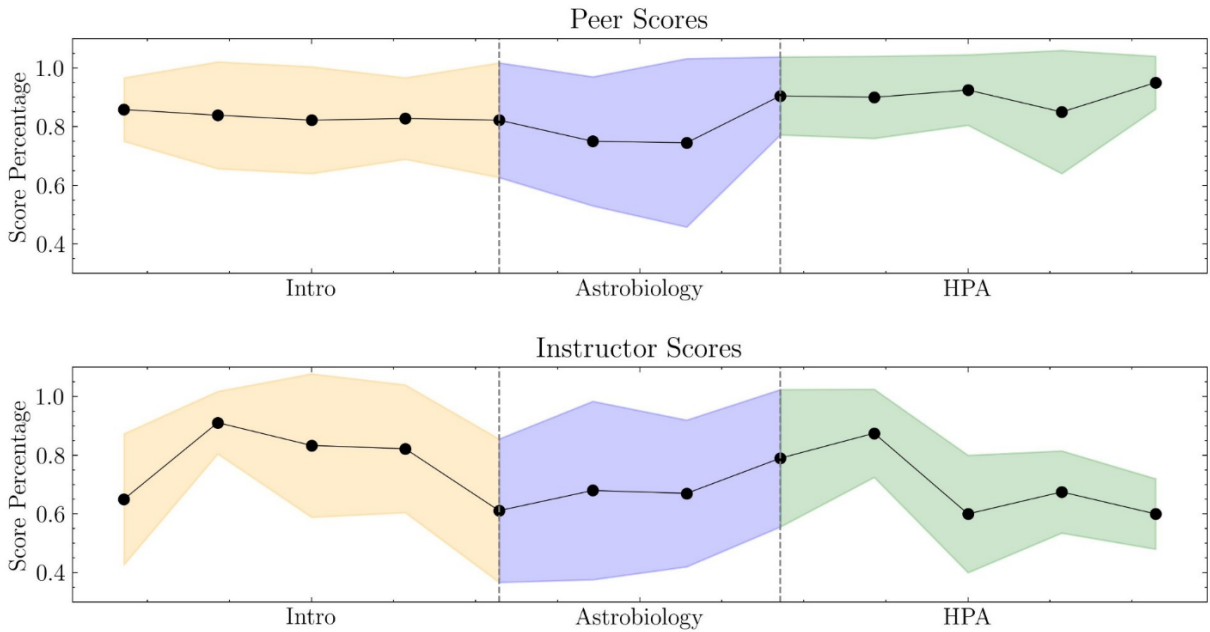

**Fig. 3** Dispersion of grades as a function of the questions in each of the three courses. The two panels show the mean scores as points, and the shaded regions (using the same color scheme as Figure 2) denote the average magnitudes of deviations of data from a question's mean. These scores are normalized to show percentages.

The peer dispersion ranges from 10.8% to 19.6%, and the instructor dispersion ranges from 10.7% to 24.4% for the introductory course. Peer dispersion ranges from 13.3% to 28.7%, and the instructor dispersion ranges from 23.4% to 30.4% for the astrobiology course. Peer dispersion ranges from 9% to 21%, and the instructor dispersion ranges from 12% to 20% for the history and philosophy course. The instructor dispersion is smaller than peer dispersion for the history and philosophy course and the astrobiology course, whereas the peer dispersion is smaller than the instructor dispersion for the introductory course. This analysis of grades for individual students affirms that GPT-4 is superior to Coursera's peer grading mechanism, and it comes very close to matching the grades assigned by the instructor.

## Discussion and Conclusions

The promise and peril of AI for education cannot be fully elucidated in a simple pilot study like this. However, the results of using an LLM to grade student writing assignments in these MOOCs are promising.

- For research question 1: Can the LLM generate grades comparable to those of an instructor? The answer is yes. GPT-4 was able to produce comparable grades to an instructor when prompted with appropriate information which, in this case, included an example answer and a rubric.
- For research question 2: Can the LLM match or exceed the reliability of peer/instructor grading? The answer is yes, GPT-4's grading is consistent and reliable, matching or exceeding the reliability of peer grading.
- For research question 3: Can the LLM create a grading rubric that will produce LLM grades comparable to that of an instructor? The answer is yes, GPT-4 was able to produce comparable grades to an instructor with LLM generated rubrics, indicating the LLM-generated rubrics were of similar utility to the instructor-provided rubrics in terms of using them for this automated grading procedure.

The performance of the LLM in matching the instructor grades is better for the astronomy and astrobiology courses than it is for the history and philosophy class, where questions are more open-ended and it is challenging even for an instructor to create a concrete rubric. In an astronomy MOOC where a direct comparison can be made, GPT-4 performed as well as peer grading in matching the instructor grades with a small dispersion. When the LLM is used to create a rubric, that prompt alone does not give good results, but it does in combination with an instructor's model answer. Lastly, these AI methods can easily be scaled to evaluate the science writing of thousands or tens of thousands of online learners in real-time. In this analysis, we have treated instructor grading as the "gold standard" and shown that GPT-4 comes close to the instructor score in three different MOOCs. Equipped with a rubric and a detailed explanation of the instructional goals for the assignment, it is likely that a future LLM could match the performance and reliability of a human instructor. In fact, since we have assumed instructors to be perfect, when in fact they are fallible, it's plausible that an LLM could one day eclipse the reliability of an instructor.

Results from this initial study have provided insights into the strengths and weaknesses of an LLM grading system. The writing assignments analyzed for this project were gathered from an existing course that was developed without any plan to grade assignments using an LLM. Our future work will explore the opportunities and challenges of applying these computer-aided systems to assessing student writing assignments. In particular, we want to examine and compare the reasoning that instructors and LLM's use to explain their grading decisions. Progress could be made in using LLMs to generate model answers and rubrics for open-ended questions and questions asking for speculation. Another approach involves the development of writing assignments and grading systems specifically designed to play to the strengths of the LLM. From an educational perspective, the kinds of student assignments that are best suited to LLM grading occupy the first five levels of Bloom's taxonomy (Krathwohl, 2002). They are: Remember, Understand, Apply, Analyze, and Evaluate. These levels all require factual knowledge, conceptual knowledge, and procedural knowledge, which can be drawn from an existing base of information on which the LLM can be or may have already been trained. Any assignments that fall in the sixth level (Create) will be more challenging for the LLM to assess, as was seen in the results from the Astrobiology class assignments. Current LLMs fall short of being able to evaluate creative assignments and any writing that requires metacognitive thinking.

## Limitations and Future Work

Although these results are promising, several limitations are worth mentioning. Even though the authors took steps to reduce selection bias by de-identifying and blinding the data, with the relatively small sample size and the use of purposeful instead of random sampling, it is possible that systematic error could have been introduced unintentionally into the results through the selection of writing assignments that are not representative of student work in some way. The grading rubrics are also a potential limitation. Although the instructors attempted to develop rubrics that produced reliable results, it is possible that the interpretation of the grading system differed between the instructor, peer-graders, and GPT-4. It is also important to note that prior researchers have found that it is still possible to fool automated systems (Filighera et al. 2020), and this study does not address these concerns. Finally, this study was conducted using data from MOOC learners. Although the content of the class is adapted from undergraduate science classes, the motivations of MOOC and undergraduate students are different, as is the necessity of accurate grading. In a free, non-credit MOOC the stakes are low. It is more important to consider how to deal with potentially inaccurate grades in a high-stakes environment (Azad et al. 2020).

This study dealt with science writing by lifelong, adult learners in MOOCs. A natural next step for this research is to apply it in the college classroom. College instructors agree that writing is an important tool for helping undergraduates learn science and apply the principles of scientific thinking (Moon, Gear, & Schultz, 2018). However, in the large introductory classes where most students get their only experience of science, the grading burden of evaluating student writing is severe. Peer grading can be used, but as in the MOOCs described here, undergraduates are not always reliable graders (Biango-Daniels & Sarvary, 2020). We plan to use LLMs to help instructors grade student writing in large General Education classes that satisfy the science requirement for non-science majors to graduate. Initially, it would provide feedback for formative assessment, where the LLM delivers a grade based on the instructor's model answer and rubric, as in this study. Beyond that, the LLM will provide reasoning for the grade and this feedback could use the claim, evidence, reasoning framework that is widely used in middle and high school science classrooms, and recently at the college level (Eden, 2023). LLMs have recently been used for fact-checking and for identifying claim-evidence pairs in scientific content (Koneru, Wu, & Rajtmajer, 2023; Wang et al., 2023; Zeng & Zubiaga, 2024). The hope is that LLMs could provide instructors and their students with assessments of the scientific validity of student writing, aiming for the "gold standard" of conceptual learning (Gere et al., 2019).

## Acknowledgements

We acknowledge many fruitful conversations with Sanlyn Buxner about the statistical analysis and educational implications of this research, and with Alexander Danehy on the computational basis for the analysis and on the strengths and weaknesses of the current generation of large language models. This work was supported in part by a grant from the National Science Foundation, award DUE-2020784.

## Statements and Declarations

### Funding

This work was supported in part by a grant from the National Science Foundation, award DUE-2020784.

### Competing Interests

The authors have no relevant financial or non-financial interests to disclose.

### Author Contributions

All authors contributed to the design and conception of the study. Evaluation material for the MOOCs was created by Matthew Wenger, who also acted as the model instructor for this project. Implementation of the LLMs for grading student writing was done by Nikhil Garuda and Shahriar Golchin. Gathering of the peer grading information was carried out by Nikhil Garuda and Sarah Stamer. The first draft of the manuscript was written entirely by Chris Impey. All authors commented on successive versions of the manuscript and all authors read and approved the final manuscript.

# Appendix

## *Appendix A.  Questions and grading rubrics*

### <u>Astronomy: Exploring Time and Space (ETS)</u>

<u>ETS Question 1:</u>

Part 1: In terms of the scientific method, how does astronomy differ from a lab science like chemistry or biology? How can astronomers be confident of their understanding of objects that are remote from the Earth?

Rubric 1: Does the writer answer the question correctly AND express their thoughts clearly using supporting facts and relevant examples? Correct answers should include a discussion of the observational rather than the experimental nature of astronomy as well as the challenges of gathering data meaning that most data is gathered from a distance or from indirect evidence.

3 points: The writer answers the question correctly AND the writer expresses their thoughts clearly, and uses supporting facts and relevant examples. Correct answers should include a discussion of the observational rather than the experimental nature of astronomy as well as the challenges of gathering data meaning that most data is gathered from a distance or from indirect evidence.
2 points: The writer answers the question correctly, but does not explain it well, use sufficient evidence, or include both of the above components.
0 points: The writer does not answer the question correctly

Part 2: Ancient cultures built some impressive structures that incorporated astronomical functions and information (Stonehenge, Chichen Itza, the Great Pyramid). A friend or acquaintance of yours tries to argue that some of these structures and artifacts are evidence of "ancient astronauts" or visits by intelligent aliens. How would you rebut or argue against this idea?

Rubric 2: Does the writer answer the question correctly AND express their thoughts clearly using supporting facts and relevant examples? Correct answers should include a discussion of how ancient technology was well within the capability of humans at the time as well as a discussion of the difficulty of interstellar travel or a lack of evidence for alien or extraterrestrial visitation.

3 points: The writer answers the question correctly AND the writer expresses their thoughts clearly, and uses supporting facts and relevant examples. Correct answers should include a discussion of how ancient technology was well within the capability of humans at the time as well as a discussion of the difficulty of interstellar travel or a lack of evidence for alien or extraterrestrial visitation.
2 points: The writer answers the question correctly, but does not explain it well, use sufficient evidence, or include both of the above components.
0 points: The writer does not answer the question correctly

<u>ETS Question 2:</u>

Part 1: What are the advantages of large telescopes? Provide at least one.

Rubric 1: Does the writer answer the question correctly AND express their thoughts clearly using supporting facts and relevant examples?

3 points: The writer answers the question correctly AND the writer expresses their thoughts clearly, and uses supporting facts and relevant examples
2 points: The writer answers the question correctly, but does not explain it well.
0 points: The writer does not answer the question correctly

Part 2: Why do astronomers want telescopes in space when putting them there is expensive?

Rubric Part 2: Does the writer answer the question correctly AND express their thoughts clearly using supporting facts and relevant examples?

3 points: The writer answers the question correctly AND the writer expresses their thoughts clearly, and uses supporting facts and relevant examples
2 points: The writer answers the question correctly, but does not explain it well.
0 points: The writer does not answer the question correctly

Part 3: What are some examples of wavelength regions beyond the spectrum of visible light where astronomers can learn about the universe? Provide at least two.

Rubric Part 3: Does the writer answer the question correctly AND the writer expresses their thoughts clearly using supporting facts and relevant examples?

3 points: The writer answers the question correctly AND the writer expresses their thoughts clearly, and uses supporting facts and relevant examples
2 points: The writer answers the question correctly, but does not explain it well. The write only includes one wavelength instead of two.
0 points: The writer does not answer the question correctly

ETS Question 3:

Part 1: What are the two main, indirect methods for finding exoplanets?

Rubric Part 1: Does the writer answer the question correctly AND express their thoughts clearly using supporting facts and relevant examples?

3 points: The writer answers the question correctly AND the writer expresses their thoughts clearly, and uses supporting facts and relevant examples
2 points: The writer answers the question correctly, but does not explain it well, use sufficient evidence, or include both of the above components.
0 points: The writer does not answer the question correctly

Part 2: Why is it so difficult to see exoplanets directly in an image?

Rubric Part 2: Does the writer answer the question correctly AND express their thoughts clearly using supporting facts and relevant examples?

3 points: The writer answers the question correctly AND the writer expresses their thoughts clearly, and uses supporting facts and relevant examples
2 points: The writer answers the question correctly, but does not explain it well.
0 points: The writer does not answer the question correctly

Part 3: What are some similarities or differences between our Solar System and new, distant planet systems? Provide at least one similarity and/or difference.

Rubric Part 3: Does the writer answer the question correctly AND express their thoughts clearly using supporting facts and relevant examples?

3 points: The writer correctly answers the question correctly AND the writer expresses their thoughts clearly, and uses supporting facts and relevant examples
2 points: The writer answers the question correctly, but does not explain it well.
0 points: The writer does not answer the question correctly

ETS Question 4:

Part 1: What is the source or cause of the Sun's light, and how do all the elements in the periodic table get produced?

Rubric Part 1: Does the writer answer the question correctly AND express their thoughts clearly using supporting facts and relevant examples?

3 points: The writer answers the question correctly AND the writer expresses their thoughts clearly, and uses supporting facts and relevant examples
2 points: The writer answers the question correctly, but does not explain it well. The write only includes one answer instead of two.
0 points: The writer does not answer the question correctly

Part 2: What is the general process by which a large diffuse cloud of gas turns into a star and surrounding planets?

Rubric Part 2: Does the writer answer the question correctly AND express their thoughts clearly using supporting facts and relevant examples?

3 points: The writer answers the question correctly AND the writer expresses their thoughts clearly, and uses supporting facts and relevant examples. Writer must mention gravity to receive full credit.
2 points: The writer answers the question correctly, but does not explain it well.
0 points: The writer does not answer the question correctly

Part 3: Name of the two end states of stars much more massive than the Sun and describe their physical properties?

Rubric Part 3: Does the writer answer the question correctly AND express their thoughts clearly using supporting facts and relevant examples?

3 points: The writer answers the question correctly AND the writer expresses their thoughts clearly, and uses supporting facts and relevant examples
2 points: The writer answers the question correctly, but does not explain it well.
0 points: The writer does not answer the question correctly

Part 1: Why do astronomers often say that large telescopes are like time machines, or equivalently, why is distant light old light?

Rubric Part 1: Does the writer answer the question correctly AND express their thoughts clearly using supporting facts and relevant examples?

3 points: The writer answers the question correctly AND the writer expresses their thoughts clearly, and uses supporting facts and relevant examples
2 points: The writer answers the question correctly, but does not explain it well.
0 points: The writer does not answer the question correctly

Part 2: What is the evidence that the universe began in a hot, dense state 13.8 billion years ago?

Rubric 2: Does the writer answer the question correctly AND express their thoughts clearly using supporting facts and relevant examples?

3 points: The writer answers the question correctly AND the writer expresses their thoughts clearly, and uses supporting facts and relevant examples.
2 points: The writer answers the question correctly, but does not explain it well.
0 points: The writer does not answer the question correctly

Part 3: The atoms in our bodies and in all the stars in all 100 galaxies form a small percentage of the contents of the universe. What are the two dominant ingredients of the universe and why are astronomers so unsure of their physical nature?

Rubric Part 3: Does the writer answer the question correctly AND express their thoughts clearly using supporting facts and relevant examples?

3 points: The writer answers the question correctly AND the writer expresses their thoughts clearly, and uses supporting facts and relevant examples.
2 points: The writer answers the question correctly, but does not explain it well.
0 points: The writer does not answer the question correctly

## Astrobiology: Exploring Other Worlds (ABIO)

ABIO Question 1:

Part 1: Clearly identify the detection methods used to gather data for each exoplanet. Briefly explain how each detection method works. Correctly identify both detection methods. Clearly explain how each detection method works.

Rubric Part 1:
2 points: The writer clearly identifies physical characteristics for both exoplanets
1 point: The writer identifies physical characteristics of only one exoplanet

0 points: The writer does not clearly identify physical characteristics of either exoplanet

Part 2: Correctly identify which physical characteristics can be learned from each set of data, and explain why. Clearly identify physical characteristics for both exoplanets.

Rubric Part 2:
4 points: The writer correctly identifies both detection methods; clear explanation of how each detection method works
3 points: The writer correctly identifies both detection methods; explanation for one or both methods is unclear or incomplete
2 points: The writer correctly identifies and explains one detection method; does not identify other detection method, no explanation
1 point: The writer correctly identifies one detection method; explanation of detection method unclear or incomplete
0 points: The writer does not identify or explain either detection method

Part 3: Clearly identify one exoplanet as Earth-like.

Rubric Part 3:
1 point: Writer clearly identifies one exoplanet as Earth-like
0 points: Writer does not clearly identify one exoplanet as Earth-like

Overall Rubric Question 1:
Student writes with sufficient clarity and detail to communicate their points effectively
2 points: Writer presents arguments in clear, logical way that demonstrates understanding of concepts; correct use of scientific terms/language. Connections between concepts are well developed.
1 point: Arguments attempt to address key concepts, some gaps in logic or comprehension. Some use of scientific terms, majority used correctly.
0 points: Arguments are difficult to follow, multiple flaws in logic. Incorrect or no use of key terms. Connections between concepts not present.

The student uses data from the graphs to support their explanations.
1 point: Yes, the writer uses data to support their explanation.
0 points: No, the writer does not use sufficient data to support their explanation.

ABIO Question 2:

Part 1: Discuss how habitable zone range and spectral type are related.

Rubric Part 1:
A well-written answer will define habitable zone and discuss how habitable zone range and star spectral type are related.
2 points: The writer defines habitable zone and discusses how habitable zone range and spectral type are related.
1 point: The writer defines habitable zone or the writer discusses how habitable zone range and spectral type are related

0 points: The writer does not define habitable zone or discuss how habitable zone range and spectral type are related

Part 2: Address Yousef's statement that all three planets likely have liquid surface water because they all orbit at 1 AU. Since the Earth orbits at 1 AU from the sun, and we know Earth has liquid surface water, then these exoplanets should as well. Clearly state whether you agree or disagree with the conclusion. Explain your answer with evidence and use data to support your answer.

Rubric Part 2:
The writer addresses Yousef's statement that "all three planets likely have liquid surface water because they all orbit at 1 AU. Since the Earth orbits at 1 AU from the sun, and we know Earth has liquid surface water, then these exoplanets should as well."
3 points: The writer clearly states whether they agree or disagree with Yousef's conclusion and explains answer with evidence and uses data to support answer.
2 points: The writer states whether they agree or disagree with Yousef's conclusion and attempts explanation, however, the explanation is incomplete or poorly supported by data.
1 point: The writer states whether they agree or disagree, they do not provide explanation, and do not use data to support their conclusion.
0 points: The writer does not address Yousef's statement.

Part 3: Clearly state whether you agree or disagree with Lora's conclusion that both exoplanets 2 & 3 will have liquid water, but not exoplanet-1. The star for exoplanet-1 is spectral type A, which is too big and hot and would evaporate water on exoplanets. But exoplanet-2 and exoplanet-3 orbit around favorable spectral types G and M, therefore they likely have liquid surface water. Explain your answer with evidence and use data to support your answer.

Rubric Part 3: The writer addresses Lora's statement that "both exoplanets 2 & 3 will have liquid water, but not exoplanet-1. The star for exoplanet-1 is spectral type A, which is too big and hot and would evaporate water on exoplanets. But exoplanet-2 and exoplanet-3 orbit around favorable spectral types G and M, therefore they likely have liquid surface water."
3 points: The writer clearly states whether they agree or disagree with Lora's conclusion and explains answer with evidence and they use data to support the answer.
2 points: The writer states whether they agree or disagree with Lora's conclusion and attempts an explanation, however, the explanation is incomplete or poorly supported by data.
1 point: The writer states whether they agree or disagree, they do not provide explanation, and do not use data to support their conclusion.
0 points: The writer does not address Lora's statement.

Overall Rubric Question 2:
The answer is written with clarity and good communication.
2 points: Presents arguments in clear, logical way that demonstrates understanding of concepts and correctly use scientific terms and language. Connections between concepts are well developed.
1 point: Arguments attempt to address key concepts, some gaps in logic or comprehension. Some use of scientific terms, majority used correctly.
0 points: Arguments are difficult to follow, multiple flaws in logic. Incorrect or no use of key terms. Connections between concepts are not present.

ABIO Question 3

Part 1: Discuss why/whether it is plausible that exobiology exists at all, in terms of exoplanet type, spectral type and orbital distance. Use data from the table and the generalized geologic timeline to support your reasoning.

Rubric Part 1:
Discuss why or whether it is plausible that exobiology exists at all, in terms of exoplanet type, spectral type and orbital distance. Use data from the table and the generalized geologic timeline to support your reasoning.
4 points: Clear discussion of plausibility of exobiology, describes habitability in terms of spectral type and orbital distance; uses data table to characterize exoplanet type; discussion supported by examples from class and data from table.
3 points: Clear discussion of plausibility of exobiology; describes habitability but leaves out mention of ONE of: spectral type, orbital distance, exoplanet type; (if exoplanet type is identified) uses data table to characterize exoplanet type; discussion supported by examples from class and data from table.
2 points: Discusses plausibility of exobiology, states exoplanet type but does not give evidence to support characterization; mentions relationship between spectral type and habitability; little use of evidence to support discussion.
1 point: Vague discussion of habitability.
0 points: No discussion.

Part 2: Clearly state the geologic eon and make a strong argument for the state of exobiology, given their choice of geologic eon. Discusses whether the exobiology is unicellular, multicellular, intelligent etc. Use geologic eon or age, exoplanet type and examples from class to support argument. Present arguments in a clear, logical way that demonstrates understanding of concepts. Correctly use scientific terms or language. Connections between concepts should be well developed.

Rubric Part 2:
Discuss whether the exobiology is unicellular, multicellular, intelligent etc. and explain your reasoning. Clearly state the geologic eon that corresponds to the age of the exoplanet system. Use data from the table and the generalized geologic timeline to support your reasoning.
4 points: Clearly states geologic eon and makes a strong argument for the state of exobiology, given their choice of geologic eon. Discusses whether the exobiology is unicellular, multicellular, intelligent etc.; uses geologic eon/age, exoplanet type and examples from class to support argument.
3 points: Provides some reasoning for the state of exobiology, given their choice of geologic eon. States geologic eon. Discusses whether the exobiology is unicellular, multicellular, intelligent etc.; uses geologic eon/age, exoplanet type and examples from class to support argument.
2 points: Attempts to explain state of evolution in connection with geologic age; discusses whether the exobiology is unicellular, multicellular, intelligent etc.; mentions geologic eon but does not clearly identify/choose one for the exoplanet; use of some data/examples to support argument.
1 point: States whether the exobiology is unicellular, multicellular, intelligent etc.; does not use evidence to support conclusion; mentions geologic eon or does not clearly identify which geologic eon they have chosen; does not use data to support argument.
0 points: No discussion of exobiology or geologic eon.

Overall Rubric Question 3:
Student writes with sufficient clarity and detail to communicate their points effectively
2 points: Presents arguments in clear, logical way that demonstrates understanding of concepts; correct use of scientific terms/language. Connections between concepts are well developed.
1 point: Arguments attempt to address key concepts, some gaps in logic or comprehension. Some use of scientific terms, majority used correctly.

0 points: Arguments are difficult to follow, multiple flaws in logic. Incorrect or no use of key terms. Connections between concepts not present.

## **Knowing the Universe: History and Philosophy of Astronomy (HPA)**

HPA Question 1:

Describe the practical and philosophical importance of astronomy for humans living a nomadic lifestyle in 20,000 B.C.

Rubric Part 1:
The author clearly states the prompt they have chosen for the assignment.
0 points: No
1 point: Yes

Rubric Part 2: The author clearly responds to the prompt, makes a compelling argument, and uses supporting evidence and/or information from the class materials.
0 points: The author did not address the prompt at all and did not use supporting evidence from the course.
1 point: The author attempted to answer respond to the prompt at a basic level but it is confusing or incomplete and does not reference course information to support the claims.
2 points: The author does a reasonable job of responding to the prompt, but it suffers from one (and only one) of the following problems: 1) the response is incomplete or unclear 2) it does not reference course information to support the claims.
3 points: The author answers the question thoroughly and thoughtfully and references information from the course to support their perspective or claims.

HPA Question 2:

According to the current understanding of the universe, the cosmos has a definite beginning but an infinite future. What are the philosophical problems and implications of this?

Rubric Part 1: The author clearly states the prompt they have chosen for the assignment.
0 points: No
1 point: Yes

Rubric Part 2: The author clearly responds to the prompt, makes a compelling argument, and uses supporting evidence and/or information from the class materials.
0 points: The author did not address the prompt at all and did not use supporting evidence from the course.
1 point: The author attempted to answer respond to the prompt at a basic level but it is confusing or incomplete and does not reference course information to support the claims.
2 points: The author does a reasonable job of responding to the prompt, but it suffers from one (and only one) of the following problems: 1) the response is incomplete or unclear 2) it does not reference course information to support the claims.
3 points: The author answers the question thoroughly and thoughtfully and references information from the course to support their perspective or claims.

HPA Question 3:

According to Enlightenment philosophy, why might liberty and personal rights be connected to the pursuit of science?

Rubric Part 1: The author clearly states the prompt they have chosen for the assignment.
0 points: No
1 point: Yes

Rubric Part 2: The author clearly responds to the prompt, makes a compelling argument, and uses supporting evidence and/or information from the class materials.
0 points: The author did not address the prompt at all and did not use supporting evidence from the course.
1 point: The author attempted to answer respond to the prompt at a basic level but it is confusing or incomplete and does not reference course information to support the claims.
2 points: The author does a reasonable job of responding to the prompt, but it suffers from one (and only one) of the following problems: 1) the response is incomplete or unclear 2) it does not reference course information to support the claims.
3 points: The author answers the question thoroughly and thoughtfully and references information from the course to support their perspective or claims.

HPA Question 4:

Imagine if the universe was eternal and unchanging as proposed by the steady state model in the 1950's. What are the philosophical and scientific implications of this?

Rubric Part 1: The author clearly states the prompt they have chosen for the assignment.
0 points: No
1 point: Yes

Rubric Part 2: The author clearly responds to the prompt, makes a compelling argument, and uses supporting evidence and/or information from the class materials.
0 points: The author did not address the prompt at all and did not use supporting evidence from the course.
1 point: The author attempted to answer respond to the prompt at a basic level but it is confusing or incomplete and does not reference course information to support the claims.
2 points: The author does a reasonable job of responding to the prompt, but it suffers from one (and only one) of the following problems: 1) the response is incomplete or unclear 2) it does not reference course information to support the claims.
3 points: The author answers the question thoroughly and thoughtfully and references information from the course to support their perspective or claims.

*Appendix B.  Example Rubrics Generated for <u>Knowing the Universe: History and Philosophy of Astronomy (HPA)</u> by GPT-4*

Rubric for HPA Question 1:

- Understanding of the topic (1 point): The student demonstrates a clear understanding of the practical and philosophical importance of astronomy for humans living a nomadic lifestyle in 20,000 B.C.
- Use of course topics and sources (1 point): The student effectively uses course topics and sources to support their claims.
- Clarity and completeness of the answer (1 point): The student's answer is clear, concise, and fully addresses the prompt.
- Originality and creativity (1 point): The student's answer is original and demonstrates creative thinking.

Rubric for HPA Question 2:

- Understanding of the topic (1 point): The student demonstrates a clear understanding of the philosophical problems and implications of the universe having a definite beginning but an infinite future.
- Use of course topics and sources (1 point): The student effectively uses course topics and sources to support their claims.
- Clarity and completeness of the answer (1 point): The student's answer is clear, concise, and fully addresses the prompt.
- Originality and creativity (1 point): The student's answer is original and demonstrates creative thinking.

Rubric for HPA Question 3:

- Understanding of the topic (1 point): The student demonstrates a clear understanding of why liberty and personal rights might be connected to the pursuit of science according to Enlightenment philosophy.
- Use of course topics and sources (1 point): The student effectively uses course topics and sources to support their claims.
- Clarity and completeness of the answer (1 point): The student's answer is clear, concise, and fully addresses the prompt.
- Originality and creativity (1 point): The student's answer is original and demonstrates creative thinking.

Rubric for HPA Question 4:

- Understanding of the topic (1 point): The student demonstrates a clear understanding of the philosophical and scientific implications of the universe being eternal and unchanging as proposed by the steady state model in the 1950's.
- Use of course topics and sources (1 point): The student effectively uses course topics and sources to support their claims.
- Clarity and completeness of the answer (1 point): The student's answer is clear, concise, and fully addresses the prompt.
- Originality and creativity (1 point): The student's answer is original and demonstrates creative thinking.

*Appendix C.  Example LLM Prompts*

**Instruction:** You are a fair and knowledgeable instructor whose task is to evaluate the student's assignments in accordance with the correct answers to each of the questions that are presented in the section that follows. Make sure you speak your thoughts aloud so that the students can understand the rationale for any points deducted. Let's grade assignments step by step.
- - -

**Questions/Answers:**

**Questions:**

What are the advantages of large telescopes? Provide at least one.

Why do astronomers want telescopes in space when putting them there is expensive?

What are some examples of wavelength regions beyond the spectrum of visible light where astronomers can learn about the universe? Provide at least two.

**Answers:**

Large telescopes collect more light and so permit fainter and more distant objects to be seen. Also, large telescopes also generally have higher angular resolution, although realizing this depends on being able to correct for blurring by the Earth's atmosphere. Telescopes in space are immune from the blurring effect of the Earth's atmosphere, which reduces the depth of observation and limits the ability to resolve or separate object close on the sky. For some wavelength regimes beyond the visible spectrum, astronomy must be done from space because those wavelengths do not penetrate the Earth's atmosphere. Telescopes in space are also immune from light pollution on Earth. Wavelength ranges include: radio, microwave, infrared, ultraviolet, X-ray, and gamma ray.

**Grade: 9/9**
- - -

**Student's Answers:**

Astronomers can learn about the universe by investigating certain wavelengths such as cosmic and gamma rays, X-rays, infrared, and radio waves. For example, the Cosmic Background Radiation (CMB) can tell astronomers something about the recombination epoch. It is very important for observational cosmology.

X-rays are produced in the universe when matter is heated to millions of degrees. These are therefore used to detect areas with high magnetic fields and extreme gravity plus areas where huge cosmic explosions occur.

Likewise, gamma ray bursts (GRBs) occur when a massive star collapses to form a neutron star or black hole. [...]

*[The answer is partially cut due to space constraints.]*

**Grade:**
- - -

7/9

The student's answer is quite comprehensive and shows a good understanding of the subject matter. However, the student did not directly answer the first question about the advantages of large telescopes. While they did mention that "The light-gathering power and the resolution of a telescope increase with increasing aperture", they did not explicitly state that these are advantages of large telescopes.

The student did a good job explaining why astronomers want telescopes in space, mentioning that they "give ultra-sharp optical images and to detect long and short wavelengths that cannot penetrate the Earth's atmosphere."

The student also did not provide specific examples of wavelength regions beyond the spectrum of visible light where astronomers can learn about the universe, which was asked in the third question. They mentioned the "peeling back of the electromagnetic spectrum, revealing for the first time the invisible universe" but did not provide specific examples as requested.

Therefore, 2 points are deducted for not directly answering the first question and not providing specific examples for the third question.

**Fig. 4** The first prompt uses the correct answers supplied by the course instructor, here the instructor of the MOOC. This is then used to generate the final grades for the assignments. The prompt also embeds the students' response into the input. Here, the LLM has no guidance or context on how the grading should be done. It is only provided with the total points for each question.

**Instruction:** You are a fair and knowledgeable instructor whose task is to evaluate the student's assignments in accordance with the correct answers and provided rubric to each of the questions that are presented in the section that follows.
Make sure that the score deduction follows the provided rubric. Let's grade assignments step by step and explain the reasons behind the point deduction given the rubric.
- - -
**Rubrics:**
Does the writer answer the question correctly and express their thoughts clearly using supporting facts and relevant examples?
3 points: The writer answers the question correctly and the writer expresses their thoughts clearly, and uses supporting facts and relevant examples
2 points: The writer answers the question correctly, but does not explain it well.
0 points: The writer does not answer the question correctly

Does the writer answer the question correctly and express their thoughts clearly using supporting facts and relevant examples?
3 points: The writer answers the question correctly and the writer expresses their thoughts clearly, and uses supporting facts and relevant examples
2 points: The writer answers the question correctly, but does not explain it well.
0 points: The writer does not answer the question correctly

Does the writer answer the question correctly and the writer expresses their thoughts clearly using supporting facts and relevant examples?
3 points: The writer answers the question correctly and the writer expresses their thoughts clearly, and uses supporting facts and relevant examples
2 points: The writer answers the question correctly, but does not explain it well. The write only includes one wavelength instead of two.
0 points: The writer does not answer the question correctly

**Questions/Answers:**
*{As Figure 1.}*
**Grade: 9/9**
- - -
**Student's Answers:**
*{As Figure 1.}*
**Grade:**
- - -
6/9
The student does not directly answer the first question about the advantages of large telescopes. While they mention that telescopes improve on the light-gathering power of the eye and allow astronomers to resolve finer details, they do not specifically state these as advantages of large telescopes. This is worth 2 points instead of 3.
The student answers the second question correctly and explains well why astronomers want telescopes in space. They mention that telescopes in space can give ultra-sharp optical images and detect long and short wavelengths that cannot penetrate the Earth's atmosphere. This is worth 3 points.
The student does not directly answer the third question about examples of wavelength regions beyond the spectrum of visible light. While they mention the peeling back of the electromagnetic spectrum, they do not provide specific examples of wavelength regions. This is worth 1 point instead of 3.

**Fig. 5** The second prompt builds on the first approach by adding the instructor-provided rubric for each of the questions in addition to the correct answer. This approach aligns more closely with how an instructor actually grades using these criteria. This ensures that the scoring and deductions match the instructor's standards.

**Instruction:** Your task is to design a rubric that addresses the following questions/answers. The followings are the homework assignments for the Astrobiology course. This course is designed for undergraduate students majoring in Astronomy. The rubric must be unbiased and adaptable, capable of fairly evaluating any kind of student writing assignment.

The scoring breakdown for each question should be as follows: the score for Question 1 is 10, the score for Question 2 is 10, and the score for Question 3 is 10. Your thoughtful consideration in creating this rubric will ensure all students' work is evaluated equitably and consistently. Make sure that the rubric will provide points that are whole numbered.

Each rubric should be dedicated to each question separately.

- - -

**Questions/Answers:**

**Question 1:**
*{As Figure 1.}*

**Answer 1:**
*{As Figure 1.}*
Full Score: 10/10

**Question 2:**
Discuss how habitable zone range and spectral type are related. [...]
*{The question is partially cut due to space constraints.}*

**Answer 2:**
The habitable zone is defined as the range of distances from a star within which liquid water can exist on a planet. [...]
*{The answer is partially cut due to space constraints.}*
Full Score: 10/10

**Question 3:**
Discuss why/whether it is plausible that exobiology exists at all, in terms of exoplanet type, spectral type and orbital distance. [...]
*{The question is partially cut due to space constraints.}*

**Answer 3:**
The star around HD-127688 has a K spectral type. [...]
*{The answer is partially cut due to space constraints.}*
Full Score: 10/10

- - -

Rubric for Question 1:

- Identification of detection methods (2 points): The student correctly identifies the detection methods used for each exoplanet.

- Explanation of detection methods (2 points): The student provides a clear and accurate explanation of how each detection method works.

- Identification of physical characteristics (2 points): The student correctly identifies the physical characteristics that can be learned from each set of data.

- Explanation of physical characteristics (2 points): The student provides a clear and accurate explanation of why these physical characteristics can be learned from the data.

- Identification of Earth-like exoplanet (2 points): The student correctly identifies one exoplanet as Earth-like. [...]

*{The answer is partially cut due to space constraints.}*

**Fig. 6** In the third prompt, we approach the LLM differently by asking it to create a rubric. Here, we provide the information of the course and give it the supporting instructor answers and total scores to create a new LLM-made rubric. The reasoning behind this is that LLMs are trained on extensive amounts of data from across the Internet and thus possess comprehensive interdisciplinary knowledge. This rubric is then used instead of the instructor's rubric from prompt 2 to get the scores from the LLM.